\documentclass{article}

\usepackage[final,nonatbib]{iai_neurips_2024}

\usepackage[utf8]{inputenc} 
\usepackage[T1]{fontenc}    
\usepackage{hyperref}       
\usepackage{url}            
\usepackage{booktabs}       
\usepackage{amsfonts}       
\usepackage{nicefrac}       
\usepackage{microtype}      
\usepackage{xcolor}         
\usepackage{graphicx}
\usepackage{subcaption}

\usepackage{alphabeta} 

\title{SignAttention: On the Interpretability of Transformer Models for Sign Language Translation}

\author{%
  Pedro Alejandro Dal Bianco$^{1}$\thanks{Equal contribution.} \\
  \texttt{pdalbianco@lidi.info.unlp.edu.ar} \\
  \And
  Oscar Agustín Stanchi$^{1,2}$\footnotemark[1] \\
  \texttt{ostanchi@lidi.info.unlp.edu.ar} \\
  \And
  Facundo Manuel Quiroga$^{1,3}$ \\
  \texttt{fquiroga@lidi.info.unlp.edu.ar} \\
  \And
  Franco Ronchetti$^{1,3}$ \\
  \texttt{fronchetti@lidi.info.unlp.edu.ar} \\
  \And
  Enzo Ferrante$^{2,4}$ \\
  \texttt{eferrante@dc.uba.ar} \\ \\
  $^{1}$Instituto de Investigación en Informática LIDI, Universidad Nacional de La Plata \\
  $^{2}$Consejo Nacional de Investigaciones Científicas y Técnicas (CONICET) \\
  $^{3}$Comisión de Investigaciones Científicas de la Pcia. de Bs. As. (CIC-PBA) \\
  $^{4}$Instituto de Ciencias de la Computación, Universidad de Buenos Aires \\
}

\begin{document}

\maketitle

\begin{abstract}

This paper presents the first comprehensive interpretability analysis of a Transformer-based Sign Language Translation (SLT) model, focusing on the translation from video-based Greek Sign Language to glosses and text. Leveraging the Greek Sign Language Dataset, we examine the attention mechanisms within the model to understand how it processes and aligns visual input with sequential glosses. Our analysis reveals that the model pays attention to clusters of frames rather than individual ones, with a diagonal alignment pattern emerging between poses and glosses, which becomes less distinct as the number of glosses increases. We also explore the relative contributions of cross-attention and self-attention at each decoding step, finding that the model initially relies on video frames but shifts its focus to previously predicted tokens as the translation progresses. This work contributes to a deeper understanding of SLT models, paving the way for the development of more transparent and reliable translation systems essential for real-world applications.

\end{abstract}

\section{Introduction}

Sign Language Translation (SLT) refers to the process of converting a continuous sign language video into a corresponding written language translation, with the aim of bridging communication gaps between the deaf and hearing communities~\cite{de2023machine}. SLT is a complex task that can be roughly decomposed into three sub-problems: 1) Sign Language Segmentation (SLS): segmenting a video of continuous sign language into many individual signs; 2) Sign Language Recognition (SLR): classifying a video of a single sign; and 3) translation of the identified signs into a written language whose grammar differs from the grammar of the sign language.

This grammatical difference adds a layer of complexity for interpreting SLT models as there is no one-to-one mapping between signs and words, and the ordering might vary. Therefore, it is significantly complex to interpret a SLT model without expert knowledge on both the grammar of the signs and the target written language. Furthermore, there is no universal sign language since each country or region has its own sign language with a particular grammar. Translation from video to Sign Language Glosses -written representation of the signs-~\cite{camgoz2020sign} constitutes an intermediate approach that can help us bypass these difficulties to better understand SLT models.

In latest years, SLT models have been improved by new deep learning architectures like the Transformer networks~\cite{camgoz2020sign,yin2020better,de2021frozen,aloysius2021incorporating}, which rely mainly on the attention mechanism. This mechanism has been proven to be interpretable and directly correlated with feature importance for tasks such as Natural Language Translation \cite{vashishth2019attention}.


In this work, we present, to the best of our knowledge, the first comprehensive analysis of interpretability of a Sign Language Translation Model, by training and studying a Transformer architecture for the Greek Sign Language using the well-known Greek Sign Language Dataset~\cite{adaloglou2020comprehensive}. We  study the attention outputs within the decoder at each decoding step, which consists of the additive combination of \textit{cross-attention}, performed over the context vector generated by the encoder, and \textit{self-attention}, that operates between tokens of the target sentence. By examining these outputs, we gain insight into how the model allocates attention across different frames and target words, thus identifying which elements are most significant for SLT based on their attention scores and magnitudes. Our work not only provides new insights into the relationship between written and sign languages but also paves the way for developing more transparent and explainable translation systems, which are crucial for deploying SLT technology in real-world applications where understanding model decisions is essential for both accuracy and user trust.

\section{Related work}


Interpretability in this field is largely underexplored, and the majority of existing research focuses primarily on SLR, rather than on the broader and more complex task of SLT. Example of this are \cite{moryossef2021evaluating,raihan2024bengali,hu2023self}, where CNNs and LSTMs SLR models are interpreted through human-grounded evaluation \cite{doshi2017towards}, SHAP \cite{lundberg2017unified}, or Grad-CAM \cite{selvaraju2017grad}. As for SLT, while some studies offer qualitative examples of interpretability, we have not identified any previous research that conducts a systematic analysis of interpretability in this domain. \cite{zhang2023heterogeneous} introduces the Heterogeneous Attention Transformer model, an end-to-end encoder–decoder transformer evaluated on the PHOENIX2014T dataset, with self-attention maps used for interpretability to demonstrate an improved attention distribution across two examples. In \cite{yin2023gloss}, the Gloss Attention SLT Network is presented, also based on a transformer architecture and evaluated on the PHOENIX14T, CSL-Daily, and SP-10 \cite{yin2022mlslt} datasets, with self-attention map visualizations shown only for a single example. Finally, in \cite{tang2022gloss}, the Gloss semantic-Enhanced Network with Online Back-Translation is proposed, integrating a gloss encoder, a pose decoder, and an online reverse gloss decoder for semantic alignment. Interpretability in the PHOENIX14T dataset is examined through cross-attention mechanism visualizations, but only two examples are provided to ensure consistency between gloss and pose sequences. Instead of providing simple qualitative examples, here we present a comprehensive interpretability analysis of a Transformer-based SLT model, shedding light on the mechanisms that allow the interaction between both sign and written languages.

\section{Experimental Setup}

\subsection{Dataset}

The Greek Sign Language Dataset is a laboratory-made dataset that contains 10,290 samples composed of RGB video, their corresponding gloss representation, and the translation to written Greek language. These samples correspond to 331 different sentences repeated many times by different signers. This dataset has two main benefits over other publicly available datasets: 1) it contains the gloss representation of each sample, useful for performing interpretability analysis; 2) since this dataset was constructed for research purposes, it contains many repetitions of each sentence and no singletons or words with low frequency that may add noise to model training (a common issue in other popular on-the-wild SL datasets such as PHOENIX-2014-T~\cite{camgoz2018rwth}). As input for the STL model, we used the positional keypoints of each video, precomputed using MediaPipe~\cite{lugaresi2019mediapipe}.

\subsection{Model}

\begin{figure}[htb]
    \centering
    \includegraphics[width=.8\linewidth]{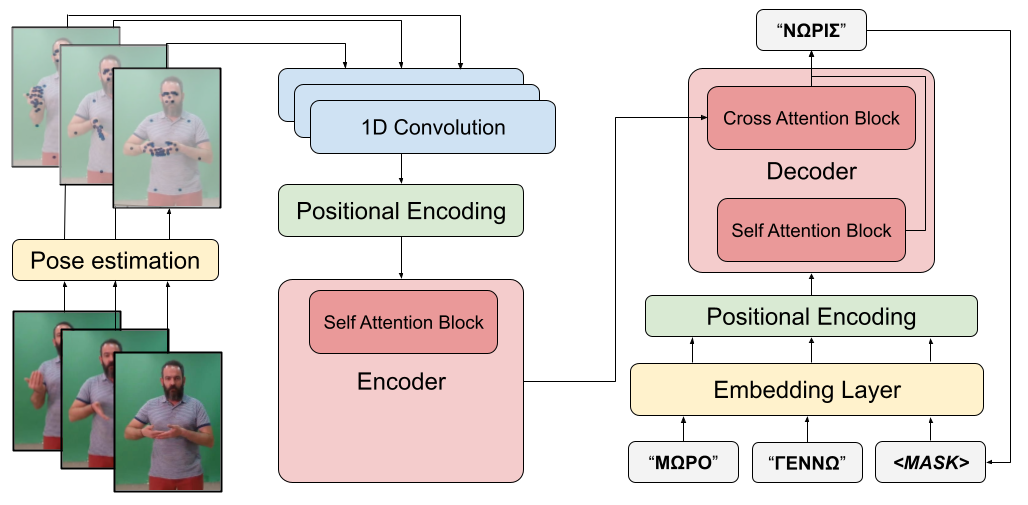}
    \caption{\textbf{Architecture of the model used from pose to gloss SLT.}}
    \label{fig:model_scheme}
\end{figure}

We trained a Transformer model~\cite{vaswani2017attention}, with a modified encoder for processing frames of pose keypoints as input data. This encoder consists of a 1 dimensional convolution run across the temporal dimension that generates an embedding of each frame. We used only one encoder layer and one decoder layer. Figure~\ref{fig:model_scheme} illustrates described model. This model and configuration were chosen as it allowed for direct interpretability of the attention weights while obtaining SoTA level Word Error Rate. For the main analysis described in this work we set the hidden dimension size to 16 to perform sign-to-gloss translation, but different sizes and sign-to-text translation were also tested. We discuss the results of these variations and their impact on the interpretability of the model in section~\ref{subsec:variations}. The results for all model configurations are shown in Table~\ref{tab:results}.

\begin{table}
  \caption{Experimental results for alternative model configurations}
  \label{tab:results}
  \centering
  \begin{tabular}{lllll}
    \toprule
    Model & Hidden Dim. Size & Target & Word Error Rate \\
    \midrule
    Ours & 16  & Gloss & 0.09  \\
    Ours & 32  & Gloss & 0.06  \\
    Ours & 64  & Gloss & 0.08  \\
    Ours & 16  & Text  & 0.1   \\
    \midrule
    SoTA & -   & Text  & 0.06  \\
    \bottomrule
  \end{tabular}
\end{table}

\section{Interpretability Analysis}


\subsection{Decoder Cross-Attention}

\begin{figure}[htb]
    \centering
    \begin{subfigure}[t]{.48\textwidth}
        \centering
        \includegraphics[width=\linewidth]{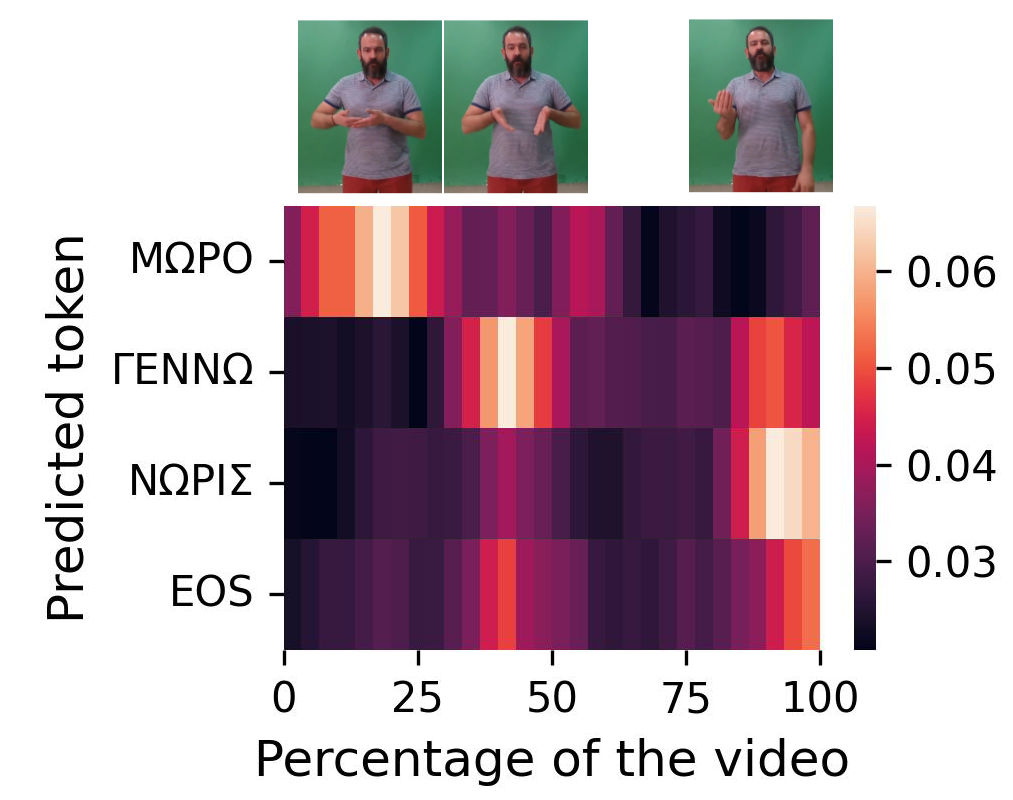}
        \caption{\textbf{Cross-attention scores across frames for each predicted token.} Above the heatmap, frames corresponding to the highest activation per gloss.}
        \label{subfig:decoder_ca_ex_1}
    \end{subfigure}\hfill
    \begin{subfigure}[t]{.48\textwidth}
        \centering
        \includegraphics[width=.8\linewidth]{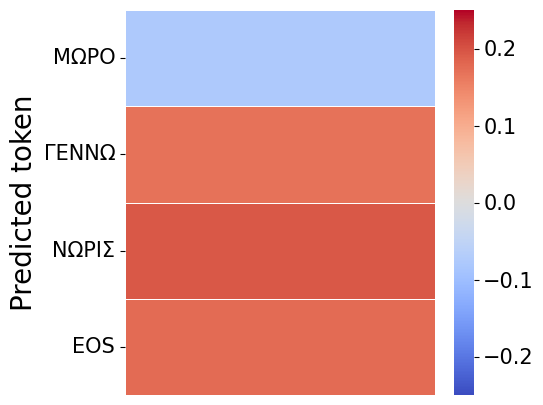}
        \caption{\textbf{Differences between self and cross-attention weights per inference.}}
        \label{subfig:sa_ca_diff_ex_1}
    \end{subfigure}
    \caption{Proposed interpretability methods applied to a single sample that translates to "ΜΩΡΟ ΓΕΝΝΩ ΝΩΡΙΣ" (BABY BIRTH EARLY).}
    \label{fig:single_sample_methods_vis}
\end{figure}

We analyzed the cross-attention scores for the encoded frames when predicting each token. A higher attention score means that more information from that particular frame was used in the prediction of a token. Figure~\ref{subfig:decoder_ca_ex_1} shows the scores obtained during the translation of a complete sentence, where we can observe the following.

\begin{enumerate}
    
    \item The model does not pay attention to individual frames but rather to clusters of contiguous frames. This behavior is not typical of the attention mechanism in text to text models, but can be observed in tasks like audio transcription where these clusters correspond to a single target token~\cite{luo2020simplified}.

    \item The attention matrix seems to be shaped as a diagonal matrix, where the position of the predicted gloss in the sentence correlates with the part of the video the model is paying attention to: for the first glosses the model pays attention to the beginning of the video and, as we move forward through the sentence, the cluster of frames to which the models is paying attention moves forward as well.
    
\end{enumerate}

As glosses have a one-to-one relationship to the signs, these results suggest that the model actually identifies these signs in the video and the corresponding mapping to glosses. Frames shown above \ref{fig:decoder_ca_ex_1}, that correspond to the highest attention scores for the prediction of each gloss, support this conclusion, as they clearly correspond to the three different signs performed.


\subsection{Self-Attention vs Cross-Attention}

In each decoding step, the cross-attention vector resulting from the previously analyzed scores for each frame, is combined with the self-attention vector by adding them to the embedding representation of each token. The vector with larger magnitude will be the one that adds more information to this new representation and thus to the prediction of the next token.

To understand the relative relevance of self and cross-attention for each prediction, we calculate the difference between the outputs of self-attention and cross-attention (averaged across the embedding dimensions). Positive values indicate a stronger contribution from the self-attention block (previous glosses), while negative values indicate a stronger contribution from the cross-attention block (the video poses).

Figure \ref{subfig:sa_ca_diff_ex_1} shows this difference for the prediction of each token of the same sentence shown in the previous example. It can be seen that as the model relies on video information for predicting the initial token, but for following predictions the self-attention has a greater influence.

\section{Global Results}

\begin{figure}[htb]
    \centering
    \begin{subfigure}[b]{1.0\textwidth}
        \centering
        \includegraphics[width=\linewidth]{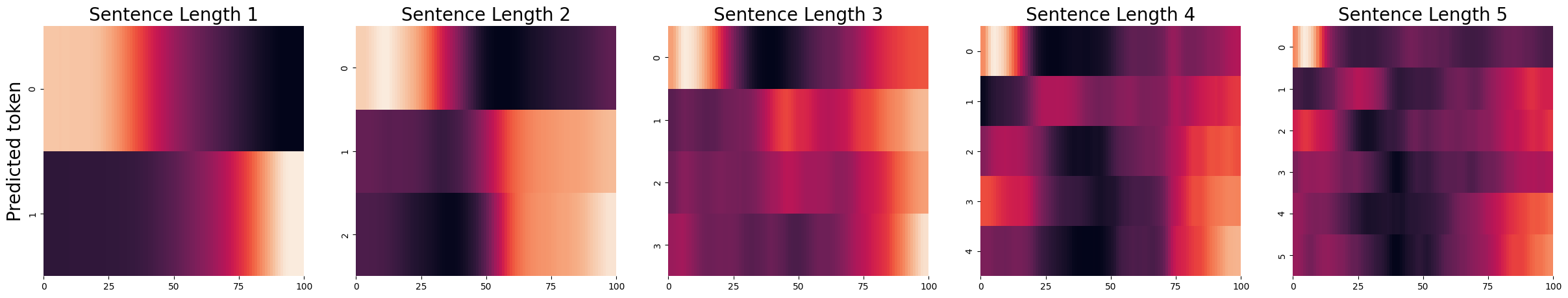}
        \caption{Cross-attention scores for sentences with 1 to 5 glosses, demonstrating a clear and consistent alignment between glosses and clusters of frames.}
        \label{subfig:avg_decoder_ca_1}
    \end{subfigure}\hfill
    \begin{subfigure}[b]{1.0\textwidth}
        \centering
        \includegraphics[width=\linewidth]{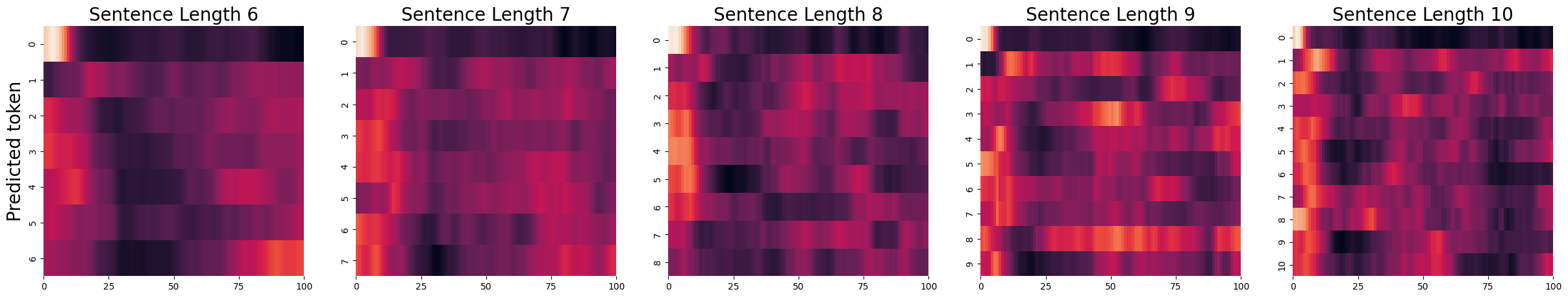}
        \caption{Cross-attention scores for sentences with 6 to 10 glosses. The alignment becomes less distinct, resulting in a more distributed attention pattern across frames.}
        \label{subfig:avg_decoder_ca_2}
    \end{subfigure}
    \caption{\textbf{Average decoder cross-attention scores for different sentence lengths.} Length of the attention weights was normalized to 100.}
    \label{fig:avg_decoder_ca}
\end{figure}

\begin{figure}[htb]
    \centering
    \includegraphics[width=\linewidth]{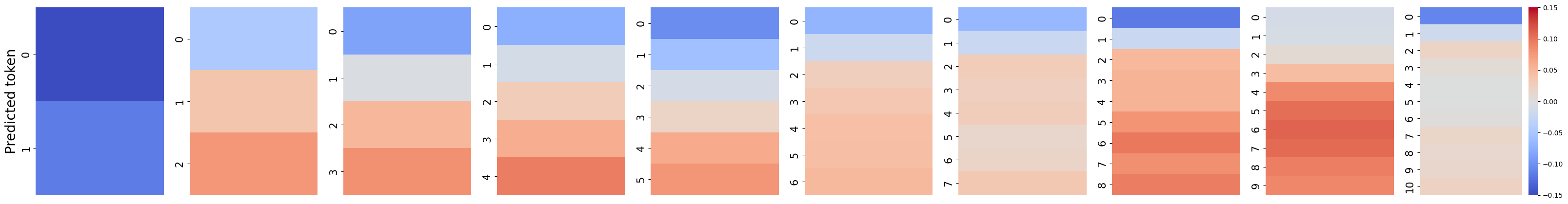}
    \caption{\textbf{Average differences between self-attention and cross-attention for different sentence lengths.}}
    \label{fig:sa_ca_diff}
\end{figure}

In the previous section, we analyzed the behavior of a single example. To further interpret the global behavior of the model in the whole dataset, we calculated cross-attention scores and the difference between self-attention and cross-attention vectors for all samples of the test set that were correctly translated (714 samples, \~81\% of the test set), as discussing a false explanation would be meaningless from an epistemological standpoint~\cite{lombrozo2016explanatory,ylikoski2010dissecting}. Then, we averaged them according to the number of glosses per sentence, which allowed us to generate an average visualization for each sentence length, shedding light into the general attention patterns for sentences with a specific number of glosses. These results are illustrated in Figure~\ref{fig:avg_decoder_ca} and Figure~\ref{fig:sa_ca_diff}.

For sentences containing up to 5 tokens, global results reveal a consistent pattern across the dataset, confirming what was observed from individual examples: 

\begin{enumerate}
    
    \item Cross-attention focuses on clusters of frames than can be mapped to the execution of a whole sign, rather than individual frames.

    \item The attention between poses and glosses aligns in a pattern resembling a diagonal matrix. This suggests that at each decoder step, the model effectively leverages the poses that correspond sequentially with the glosses, demonstrating a clear relationship between specific glosses and associated pose groups

    \item The relevance of the self-attention scores over cross-attention scores monotonically increases alongside the decoder step.
    
\end{enumerate}

However, these patterns become less pronounced as the number of glosses increases. This can be due to several reasons: As more signs are performed per video, and given that the duration of each sign per frame is variable, it is expected for later signs to appear in wider ranges of the video. Also, when more tokens are available, after initially focusing on frames, the model can rely mostly on tokens and use little to no information from the video. In effect, the model is able to predict the whole sentence without the need of identifying individual signs. The fact that in sentences with 8 and 9 glosses, only the first and second rows —where cross-attention dominates over self-attention— retain a semblance of the earlier diagonal pattern, supports   this hypothesis. Finally, it is important to highlight also that there is a significant difference in the amount and variety of examples per sequence length: 84\% of the dataset is composed of 201 different sentences of less than 6 glosses, while videos with 7 or more glosses constitute only 16\% of the dataset and contain 47 different sentences. Having a smaller set of possible target sentences might allow the model to translate the complete sentence with little information from the video, without relying on identifying individual signs.

\subsection{Experiment variants}\label{subsec:variations}

\subsubsection{Embedding size and encoder visualization}

\begin{figure}[htb]
    \centering
    \begin{subfigure}[b]{1.0\textwidth}
        \centering
        \includegraphics[width=\linewidth]{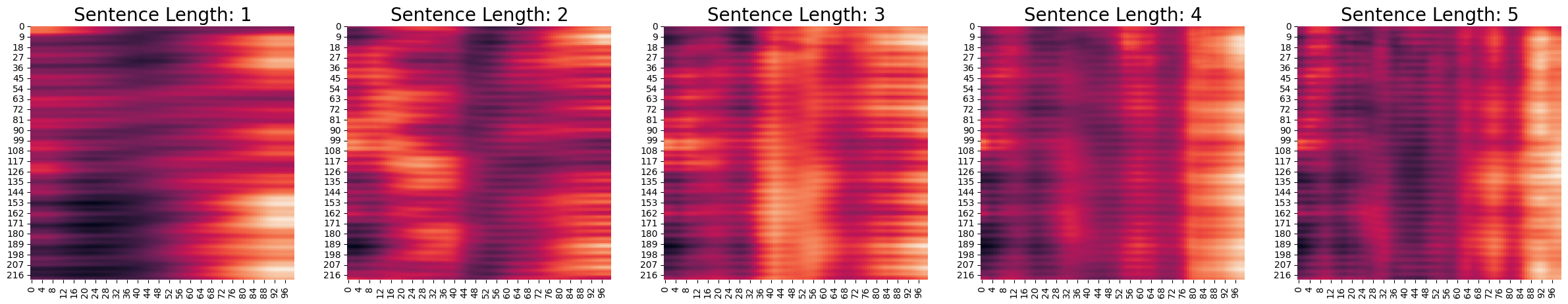}
        \caption{Model with hidden dimension set to 16. This is the same one used for obtaining all other visualizations.}
        \label{subfig:avg_encoder_sa_16}
    \end{subfigure}\hfill
    \begin{subfigure}[b]{1.0\textwidth}
        \centering
        \includegraphics[width=\linewidth]{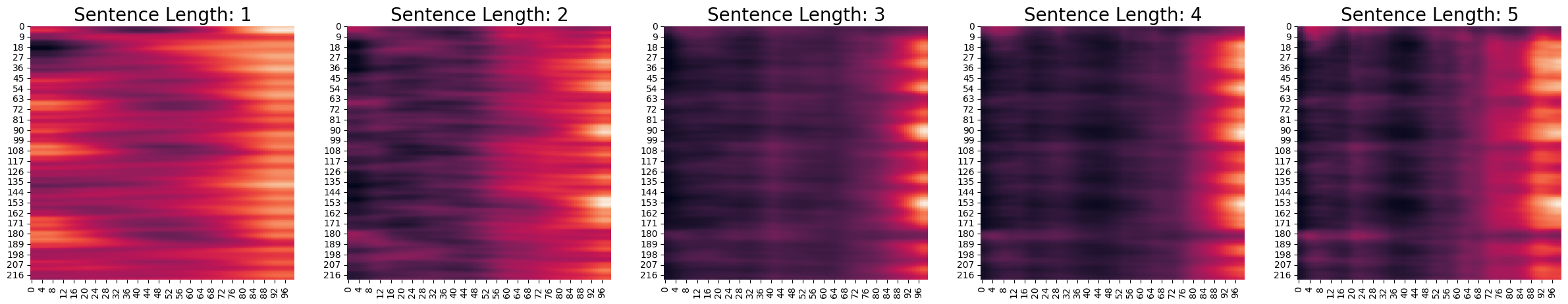}
        \caption{Model with hidden dimension set to 32.}
        \label{subfig:avg_encoder_sa_32}
    \end{subfigure}
    \caption{\textbf{Average encoder self-attention scores for different sentence lengths.} The number of frames was normalized to 100.}
    \label{fig:avg_encoder_sa}
\end{figure}

As shown in Table \ref{tab:results}, the size of the hidden dimension did not have a significant impact on its accuracy; however, it did have an impact in terms of interpretability. The patterns described in this work were not found for larger sizes of the hidden dimension. A possible explanation is that, as the hidden dimension sets the size of the representation of each frame generated by the encoder, larger representations allow the encoder to store information about the whole video in a few frames. Figure \ref{fig:avg_encoder_sa} appears to support this hypothesis. The encoder self-attention scores of the 16-dimensions model seem to be distributed along all the frames, seemingly forming as many different patterns as the number of signs in the sentences. On the other hand, for the 32-dimension model, in most cases it seems that only the last frames are being activated.

\subsubsection{Text based model}

We also performed an interpretability analysis on a model that performs proper SLT generating text instead of glosses. The decoder also learned to pay attention to clusters of frames, but these clusters were not diagonally aligned. This is expected due to the grammatical differences and the no one-to-one mapping between written Greek and Greek Sign Language. In addition, cross-attention was more relevant than self-attention for most of the sequence lengths. Figures corresponding to these results can be found in the Appendix.

\section{Conclusions and future works}

We performed an interpretability analysis of SLT models based on the Transformer architecture by generating different visualizations based on attention scores. In summary, our interpretability analysis suggests that:

\begin{enumerate}

    \item Transformer models for SLT learn to pay attention to sequencial clusters rather than individual frames.\label{item:conclusion_clusters}

    \item Cross-attention scores show a diagonal alignment between attended frames and gloss predicted, suggesting that the model is identifying the sign corresponding to each gloss.\label{item:conclusion_diag}

    \item For the translation of a whole sentence, the model initially relies more in video frames but gradually shifts its focus to the previously predicted glosses.
    
    \item When translating longer sequences, the model relies more in glosses than in video frames, and patterns described in items \ref{item:conclusion_clusters} and \ref{item:conclusion_diag} cannot be seen that clearly.

    \item We found patterns of sign segmentation in the encoder self-attention scores. These patterns are not present for bigger embedding sizes, harming the whole interpretability of the model.

    \item When trained to generate text, the model still learns to focus on clusters of frames; however, these clusters do not align diagonally with words, as expected due to the grammatical differences between the languages. Determining whether the resulting alignment is meaningful requires validation by an expert translator.

\end{enumerate}

For future work, we plan to perform a more in-depth review of the results obtained for sign-to-text translation, incorporating the view of expert translators. Moreover, we plan to establish a formal definition of the clusters of frames that the model pays attention to, so we can quantitatively evaluate their identification and alignment, aiming to use these clusters for sign segmentation. 
Finally, we aim to replicate this analysis in other SLT datasets to analyze the generalizability of the obtained results to other languages.


\medskip

{ 
\small
\bibliographystyle{nips}
\bibliography{bibliography}
}

\appendix

\section{Appendix / supplemental material}

\subsection{Other single samples visualizations}

\begin{figure}[htb]
    \centering
    \begin{subfigure}[b]{.375\textwidth}
        \centering
        \includegraphics[width=\linewidth]{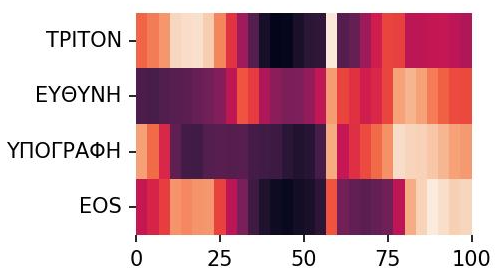}
        \caption{}
        \label{subfig:subfig1}
    \end{subfigure}\hfill
    \begin{subfigure}[b]{.625\textwidth}
        \centering
        \includegraphics[width=\linewidth]{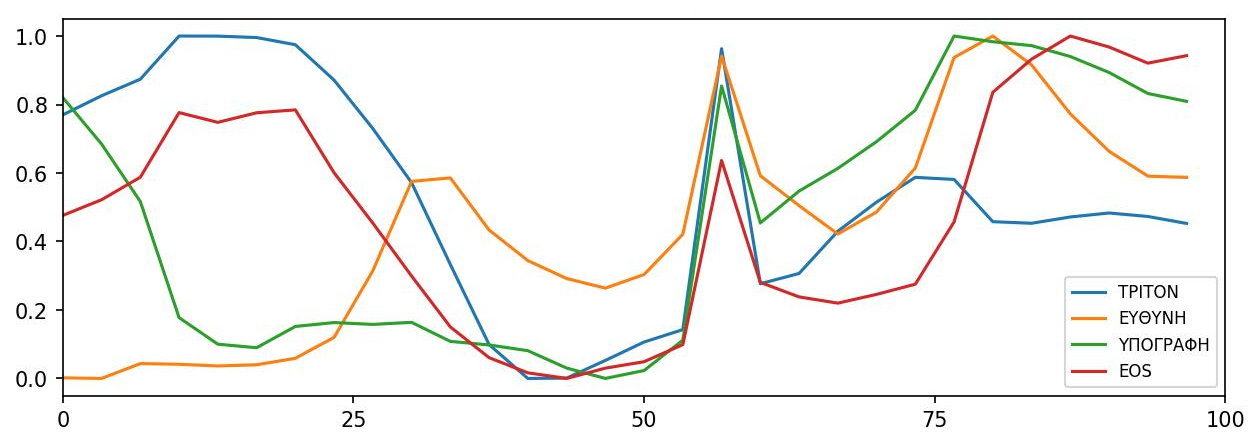}
        \caption{}
        \label{subfig:subfig2}
    \end{subfigure}
    \begin{subfigure}[b]{.375\textwidth}
        \centering
        \includegraphics[width=\linewidth]{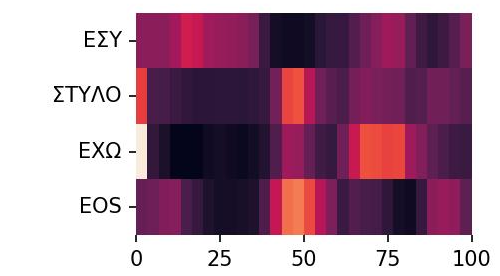}
        \caption{}
        \label{subfig:subfig3}
    \end{subfigure}\hfill
    \begin{subfigure}[b]{.625\textwidth}
        \centering
        \includegraphics[width=\linewidth]{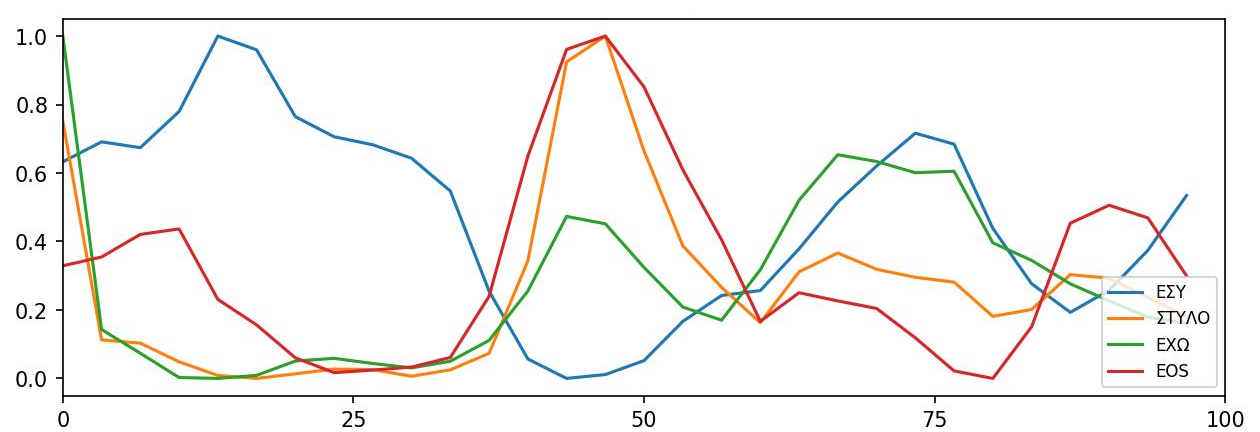}
        \caption{}
        \label{subfig:subfig4}
    \end{subfigure}
    \begin{subfigure}[b]{.375\textwidth}
        \centering
        \includegraphics[width=\linewidth]{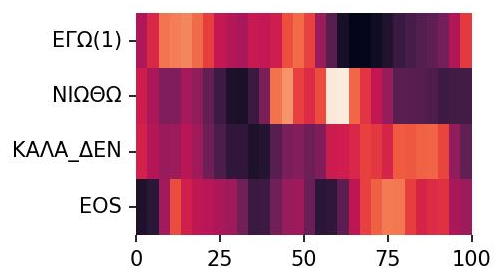}
        \caption{}
        \label{subfig:subfig5}
    \end{subfigure}\hfill
    \begin{subfigure}[b]{.625\textwidth}
        \centering
        \includegraphics[width=\linewidth]{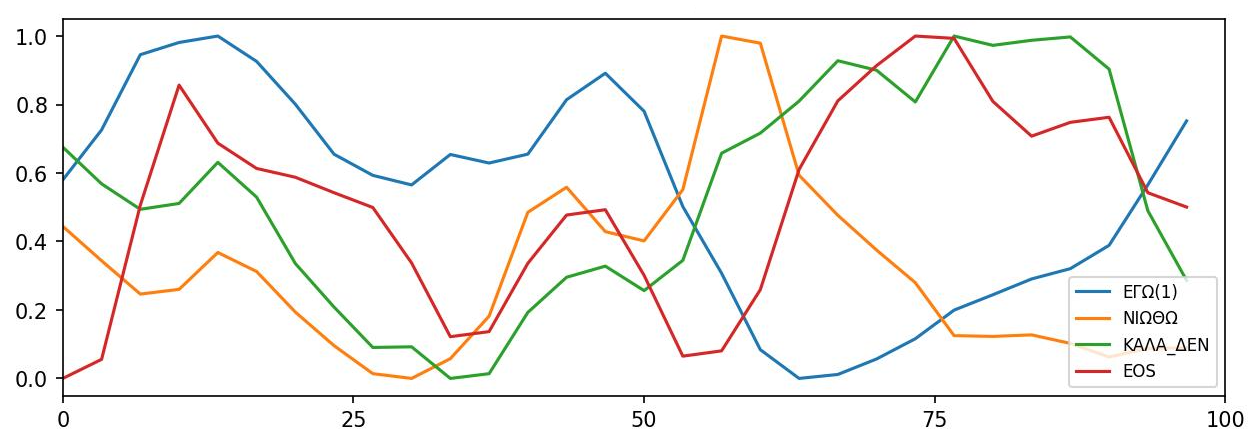}
        \caption{}
        \label{subfig:subfig6}
    \end{subfigure}
    \begin{subfigure}[b]{.375\textwidth}
        \centering
        \includegraphics[width=\linewidth]{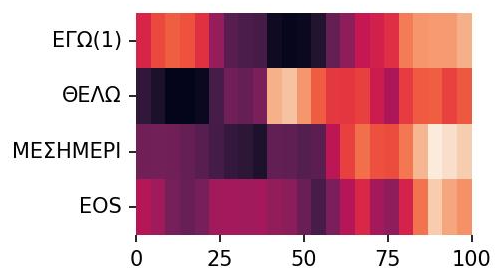}
        \caption{}
        \label{subfig:subfig7}
    \end{subfigure}\hfill
    \begin{subfigure}[b]{.625\textwidth}
        \centering
        \includegraphics[width=\linewidth]{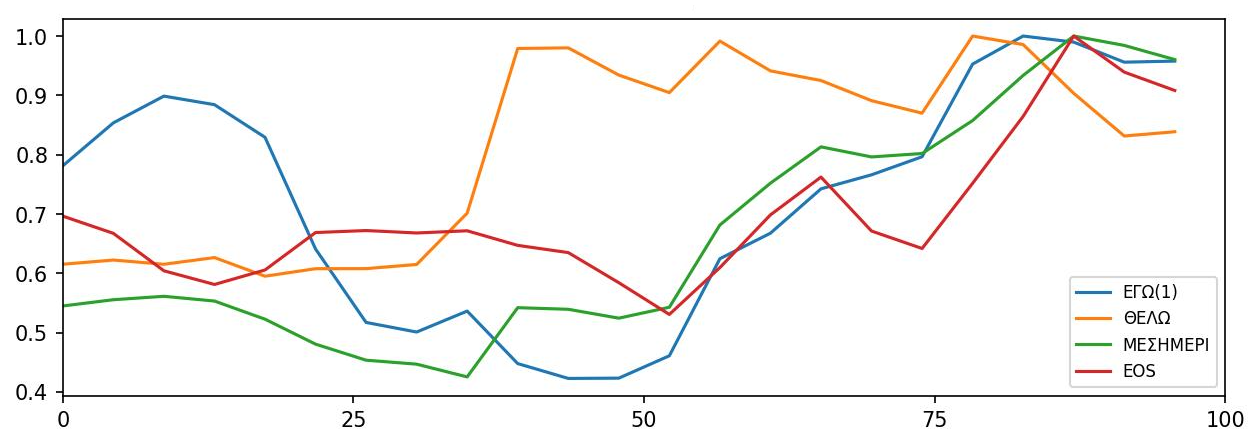}
        \caption{}
        \label{subfig:subfig8}
    \end{subfigure}
    \begin{subfigure}[b]{.44\textwidth}
        \centering
        \includegraphics[width=\linewidth]{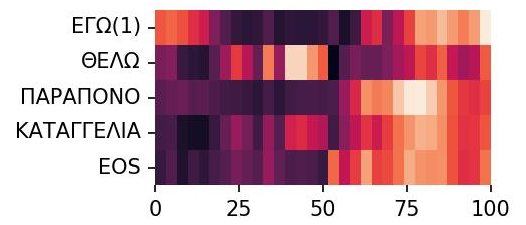}
        \caption{}
        \label{subfig:subfig9}
    \end{subfigure}\hfill
    \begin{subfigure}[b]{.56\textwidth}
        \centering
        \includegraphics[width=\linewidth]{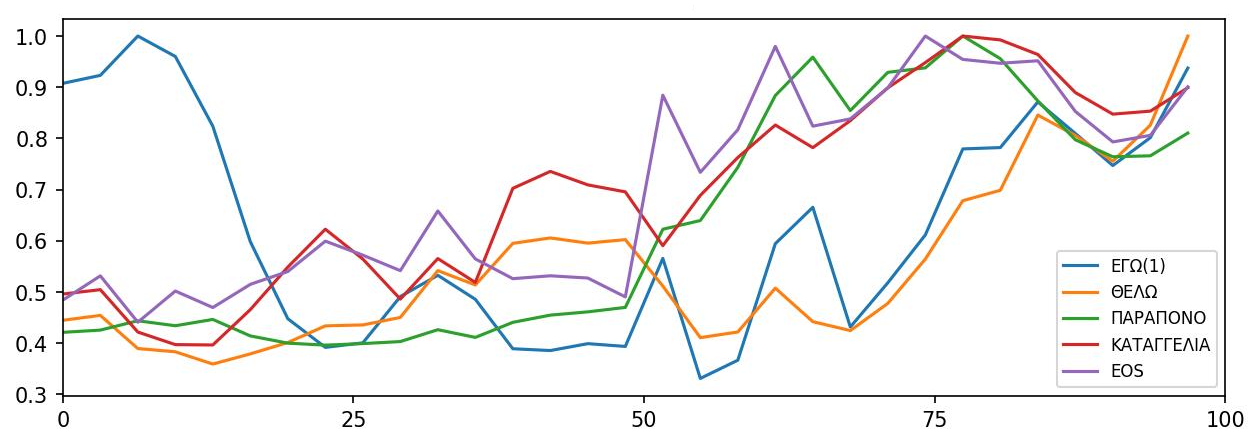}
        \caption{}
        \label{subfig:subfig10}
    \end{subfigure}
    \caption{Examples of visualization of cross-attention mechanisms in the decoder. On the left, a heatmap illustrating the cross-attention distribution across frames for each predicted token in the output sequence. On the right, a line plot representing the normalized attention weights over the video sequence. The peaks indicate the frames that the model focused on most for each token prediction.}
    \label{fig:decoder_ca_ex_1}
\end{figure}

\newpage

\subsection{Sign-to-text model results}

\begin{figure}[htb]
    \centering
    \begin{subfigure}[b]{\textwidth}
        \centering
        \includegraphics[width=\linewidth]{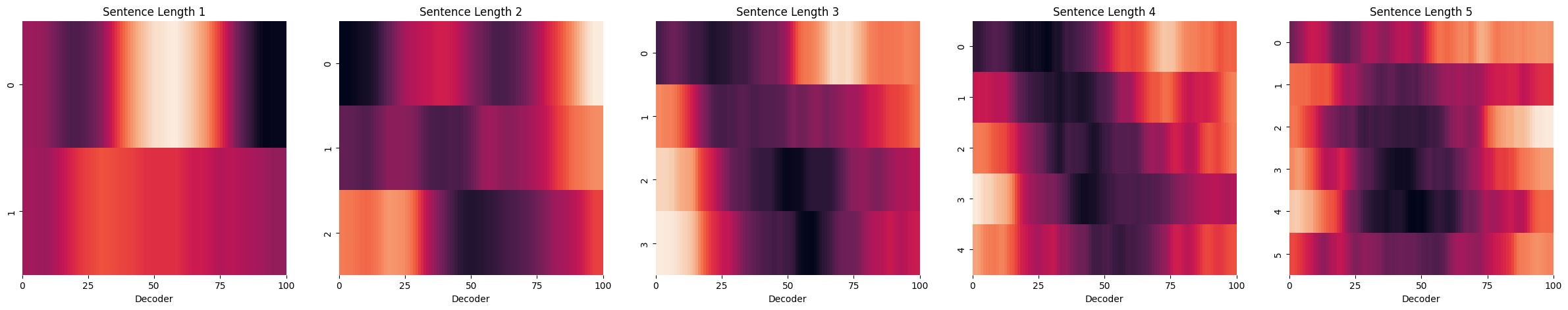}
        \caption{Average decoder cross-attention scores for sentences with 1 to 5 tokens.}
    \end{subfigure}\hfill
    \begin{subfigure}[b]{\textwidth}
        \centering
        \includegraphics[width=\linewidth]{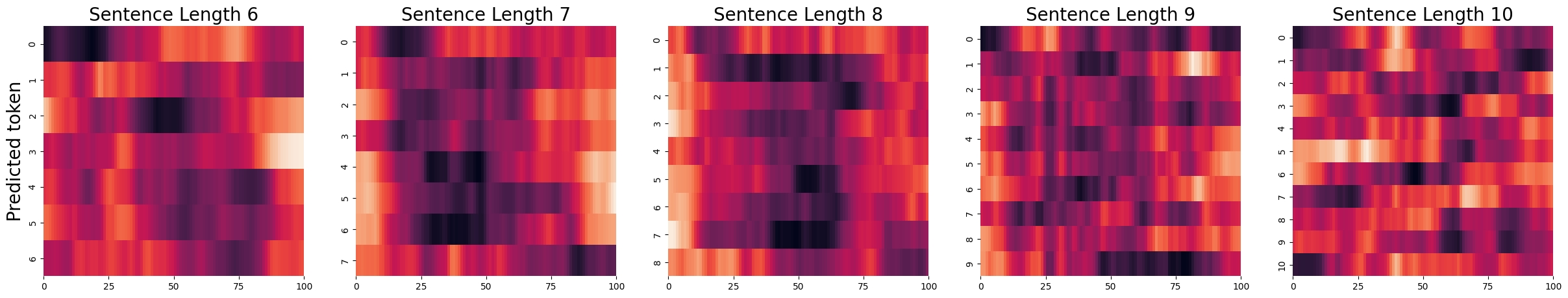}
        \caption{Average decoder cross-attention scores for sentences with 6 to 10 tokens.}
    \end{subfigure}\hfill
    \begin{subfigure}[b]{\textwidth}
        \centering
        \includegraphics[width=\linewidth]{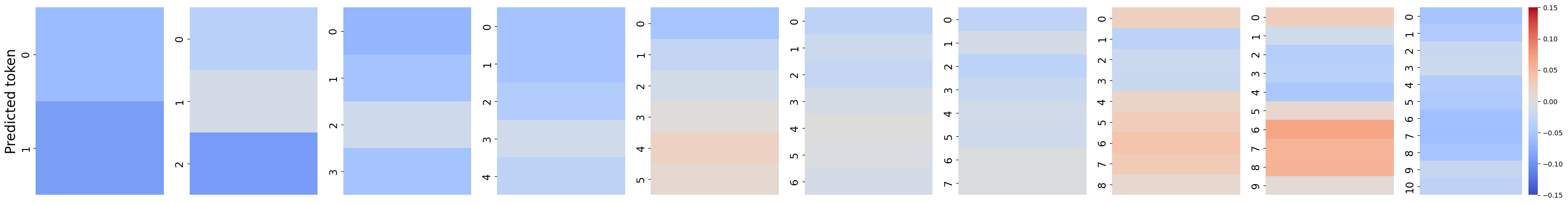}
        \caption{Average differences between self-attention and cross-attention for different sentence lengths.}
    \end{subfigure}
    \begin{subfigure}[b]{\textwidth}
        \centering
        \includegraphics[width=\linewidth]{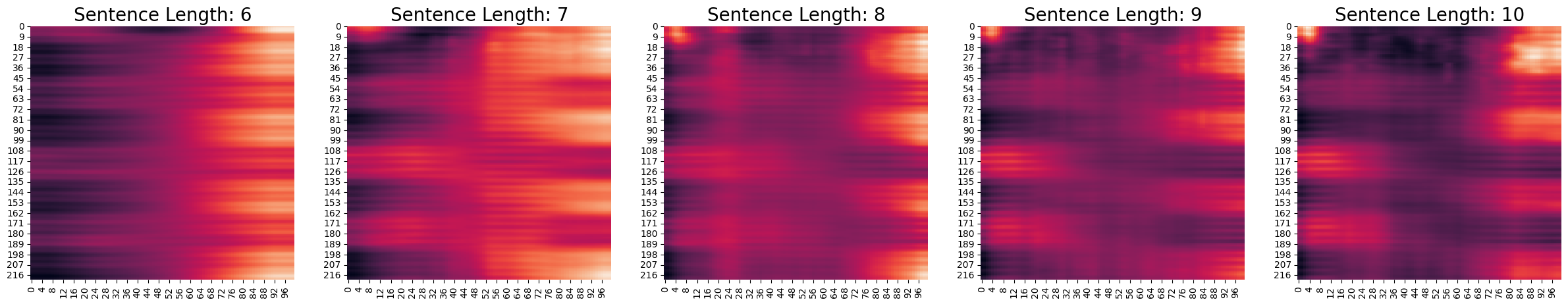}
        \caption{Average encoder self-attention scores for different sentence lengths.}
    \end{subfigure}\hfill
    \caption{Visualization of dataset-wide results for sign-to-text translation model.}
    \label{fig:text_vis}
\end{figure}

\end{document}